# Unsupervised comparable corpora preparation and exploration for bi-lingual translation equivalents

*Krzysztof Wołk, Krzysztof Marasek*

Department of Multimedia
Polish-Japanese Academy of Information Technology, Koszykowa 86, 02-008 Warsaw
kwolk@pja.edu.pl, kmarasek@pja.edu.pl

## Abstract

The multilingual nature of the world makes translation a crucial requirement today. Parallel dictionaries constructed by humans are a widely-available resource, but they are limited and do not provide enough coverage for good quality translation purposes, due to out-of-vocabulary words and neologisms. This motivates the use of statistical translation systems, which are unfortunately dependent on the quantity and quality of training data. Such systems have a very limited availability especially for some languages and very narrow text domains. In this research we present our improvements to current comparable corpora mining methodologies by re-implementation of the comparison algorithms (using Needleman-Wunch algorithm), introduction of a tuning script and computation time improvement by GPU acceleration. Experiments are carried out on bilingual data extracted from the Wikipedia, on various domains. For the Wikipedia itself, additional cross-lingual comparison heuristics were introduced. The modifications made a positive impact on the quality and quantity of mined data and on the translation quality.

## 1. Introduction

The aim of this research is a preparation of parallel and comparable corpora and language models. This work improves SMT quality through the processing and filtering of parallel corpora and through extraction of additional parallel data from the resulting comparable corpora. To enrich the language resources of SMT systems, adaptation and interpolation techniques have been applied to the prepared data. Experiments were conducted using data from a wide domain (TED[1] presentations on various topics).

Evaluation of SMT systems was performed on random samples of parallel data using automated algorithms (BLEU metric) to evaluate the quality and potential usability of the SMT systems' output [1].

As far as experiments are concerned, the Moses Statistical Machine Translation Toolkit software [2] is used. Moreover, the multi-threaded implementation of the GIZA++ tool is employed to train models on parallel data and to perform their symmetrization (using Berkeley Aligner [28]) at the phrase level. The statistical language models from single-language data are trained and smoothed using the SRI Language Modeling toolkit (SRILM). In addition, data from outside the thematic domain is adapted. In the case of parallel models, Moore-Lewis Filtering [3] is used for pseudo in-domain data selection, while single-language models are linearly interpolated [4].

Lastly, methodology proposed in the Yalign [5] parallel data mining tool is analyzed and enhanced. Its speed is increased by reimplementing it in a multi-threaded manner and by employing graphics processing unit (GPU) for its calculations. Quality is improved by using the Needleman-Wunch [6] algorithm for sequence comparison and by developing a tuning script that adjusts mining parameters to specific domain requirements.

The resulting systems out-performed baseline systems used in the tests.

## 2. Corpora Types

A corpus is a large collection of texts, stored on a computer. Text collections are called corpora. The term "parallel corpus" is typically used in linguistic circles to refer to texts that are translations of each other. For statistical machine translation, we are especially interested in parallel corpora, which are texts paired with a translation into another language. Preparing parallel texts for the purpose of statistical machine translation may require crawling the web, extracting the text from formats such as HTML, and performing document and sentence alignment [4].

There are two main types of parallel corpora, which contain texts in two languages. In a comparable corpus, the texts are of the same kind and cover the same content. An example is a corpus of articles about football from English and Polish newspapers. In a translation corpus, the texts in one language (e) are translations of texts in the second language (f). It is important to remember that the term "comparable corpora" refers to texts in two languages that are similar in content, but are not translations of each other [4].

To exploit a parallel text, some kind of text alignment, which identifies equivalent text segments (approximately, sentences), is a prerequisite for analysis.

Machine translation algorithms for translating between a first language and a second language are often trained using parallel fragments, comprising a first language corpus and a second language corpus, which is an element-for-element translation of the first language corpus. Such training may involve large training sets that may be extracted from large bodies of similar sources, such as databases of news articles written in the first and second languages describing similar events. However, extracted fragments may be comparatively "noisy", with extra elements inserted in each corpus. Extraction techniques may be devised that can differentiate between "bilingual" elements represented in both corpora and "monolingual" elements represented in only one corpus, and for extracting cleaner parallel fragments of bilingual elements. Such techniques may involve conditional probability determinations on one corpus with respect to the other corpus, or joint probability determinations that concurrently evaluate both corpora for bilingual elements [4].

[1] https://www.ted.com/

Because of such difficulties, high-quality parallel data is difficult to obtain, especially for less popular languages. Comparable corpora are the answer to the problem of lack of data for the translation systems for under-resourced languages and subject domains. It may be possible to use comparable corpora to directly obtain knowledge for translation purposes. Such data is also a valuable source of information for other cross-lingual, information-dependent tasks. Unfortunately, such data is quite rare, especially for the Polish–English language pair. On the other hand, monolingual data for those languages is accessible in far greater quantities [4].

Summing up, four main corpora types can be distinguished. Most rare parallel corpora can be defined as corpora that contain translations of the same document into two or more languages. Such data should be aligned, at least at the sentence level. A noisy parallel corpus contains bilingual sentences that are not perfectly aligned or have poor quality translations. Nevertheless, mostly bilingual translations of a specific document should be present in it. A comparable corpus is built from non-sentence-aligned and untranslated bilingual documents, but the documents should be topic-aligned. A quasi-comparable corpus includes very heterogeneous and non-parallel bilingual documents that may or may not be topic-aligned [18].

## 3. State of the art

As far as comparable corpora are concerned, many attempts (especially for Wikipedia) have been made so far to extract parallel data samples. Two main approaches for building comparable corpora can be distinguished. Perhaps the most common approach is based on the retrieval of cross-lingual information. In the second approach, source documents must be translated using any machine translation system. The documents translated in that process are then compared with documents written in the target language, to find the most similar document pairs.

An interesting idea for mining parallel data from Wikipedia was described in [8]. The authors propose two separate approaches. The first idea is to use an online machine translation (MT) system to translate Dutch Wikipedia pages into English, and then try to compare original EN pages with translated ones. The idea, although interesting, seems computationally infeasible, and it presents a chicken-egg problem. Their second approach uses a dictionary generated from Wikipedia titles and hyperlinks shared between documents. Unfortunately, the second method was reported to return numerous, noisy sentence pairs. The second method was improved in [9] by additional restrictions on the length of the correspondence between chunks of text and by introducing an additional similarity measure. They prove that [8] the precision (understood as number of correct translations pairs over total number of candidates) is about 21%, and in the improved method [9], the precision is about 43%.

Yasuda and Sumita [11] proposed an MT bootstrapping framework based on statistics that generate a sentence-aligned corpus. Sentence alignment is achieved using a bilingual lexicon that is automatically updated by the aligned sentences. Their solution uses a corpus that has already been aligned for initial training. They showed that 10% of Japanese Wikipedia sentences have an equivalent on English Wikipedia.

Interwiki links were leveraged by the approach of Tyers and Pienaar in [10]. Based on Wikipedia link structure, a bilingual dictionary is extracted. In their work, they measured the average mismatch between linked Wikipedia pages for different languages. They found that precision of their method is about 69-92% depending on language.

In [12] the authors attempt to advance the state of the art in parallel data mining by modeling document-level alignment using the observation that parallel sentences can most likely be found in close proximity. They also use annotation available on Wikipedia and an automatically-induced lexicon model. The authors report 90% recall and 80% precision.

The author of [13] introduces an automatic alignment method for parallel text fragments that uses a textual entailment technique and a phrase-based SMT system. The author states that significant improvements in SMT quality were obtained (BLEU increased by 1.73) by using this aligned data between German and French languages.

Another approach for exploring Wikipedia was recently described in [14] by M. Plamada and M. Volk. Their solution differs from the previously described methods in which the parallel data was restricted by the monotonicity constraint of the alignment algorithm used for matching candidate sentences. Their algorithm ignores the position of a candidate in the text and, instead, ranks candidates by means of customized metrics that combine different similarity criteria. In addition, the authors limit the mining process to a specific domain and analyze the semantic equivalency of extracted pairs. The mining precision in their work is 39% for parallel sentences and 26% for noisy-parallel sentences, with the remaining sentences misaligned. They also report an improvement of 0.5 points in the BLEU metric for out-of-domain data, and almost no improvement for in-domain data.

The authors in [15] propose obtaining only title and some meta-information, such as publication date and time for each document, instead of its full contents, to reduce the cost of building the comparable corpora. The cosine similarity of the titles' term frequency vectors were used to match titles and the contents of matched pairs.

In the research described in [16], the authors introduce a document similarity measure that is based on events. To count the values of this metric, they model documents as sets of events. These events are temporal and geographical expressions found in the documents. Target documents are ranked based on temporal and geographical hierarchies.

The authors in [17] also suggest an automatic technique for building a comparable corpus from the web using news web pages, Wikipedia, and Twitter in. They extract entities, time interval filtering, URLs of web pages, and document lengths as features for classification and for gathering the comparable data.

In the present research, a method inspired by the Yalign tool is used. The solution was far from perfect, but after improvements that were made during this research, it supplied the SMT systems with bi-sentences of good quality in a reasonable amount of time.

## 4. Parallel data mining

In this research, methodologies that obtain parallel corpora from data sources that are not sentence-aligned, such as noisy parallel or comparable corpora, are presented. The results of initial experiments on text samples obtained from Wikipedia pages are presented. We chose Wikipedia as a data source because of the large number of documents that it provides (4,524,017 on EN wiki, at the time of writing). Furthermore, Wikipedia contains not only comparable documents, but also some documents that are translations of each other. The quality of the approach used was measured by improvements in MT systems translations.

For the experiments in data mining, the TED corpora prepared for the IWSLT 2015 evaluation campaign by FBK[1] were chosen. This domain is very wide and covers many unrelated subject areas. The data contains almost 2.5M untokenized words [19]. The experiments were conducted on DE-EN, FR-EN, VI-EN and CS-EN corpora.

The solution can be divided into three main steps. First, the comparable data is collected, then it is aligned at the article level, and finally the aligned results are mined for parallel sentences. The last two steps are not trivial, because there are great disparities between Wikipedia documents. This is most likely why sentences in the raw Wiki corpus are mostly misaligned, with translation lines whose placement does not correspond to any text lines in the source language. Moreover, some sentences have no corresponding translations in the corpus at all. The corpus might also contain poor or indirect translations, making alignment difficult. Thus, alignment is crucial for accuracy. Sentence alignment must also be computationally feasible to be of practical use in various applications.

Before a mining tool processes the data, texts must be prepared. Firstly, all the data is saved in a relational database. Secondly, our tool aligns article pairs and removes from the database articles that appear only in one of the two languages. These topic-aligned articles are filtered to remove any HTML tags, XML tags, or noisy data (tables, references, figures, etc.). Finally, bilingual documents are tagged with a unique ID as a topic-aligned, comparable corpus. To extract the parallel sentence pairs, a decision was made to try strategy designed to automate the parallel text mining process by finding sentences that are close translation matches from comparable corpora. This presents opportunities for harvesting parallel corpora from sources, like translated documents and the web, that are not limited to a particular language pair. However, alignment models for two selected languages must first be created.

The solution was implemented using a sentence similarity metric that produces a rough estimate (a number between 0 and 1) of how likely it is for two sentences to be a translation of each other. It also uses a sequence aligner, which produces an alignment that maximizes the sum of the individual (per sentence pair) similarities between two documents [5].

For sequence alignment, the Yalign used an A* search approach [7] to find an optimal alignment between the sentences in two selected documents. The algorithm has a polynomial time worst-case complexity, and it produces an optimal alignment. Unfortunately, it cannot handle alignments that cross each other or alignments from two sentences into a single one [7].

After the alignment, only sentences that have a high probability of being translations are included in the final alignment. The result is filtered in order to deliver high quality alignments. To do this, a threshold is used: if the sentence similarity score is below it, the pair is excluded.

[1] http://www.fbk.eu/

For the sentence similarity metric, the algorithm uses a statistical classifier's likelihood output and normalizes it into the 0–1 range.

The classifier must be trained in order to determine if sentence pairs are translations of each other. A Support Vector Machine (SVM) classifier was used in this research. Besides being an excellent classifier, an SVM can provide a distance to the separation hyperplane during classification, and this distance can be easily modified using a Sigmoid Function to return a value similar to likelihood between 0 and 1 [21].

The use of a classifier means that the quality of the alignment depends not only on the input but also on the quality of the trained classifier.

To train the classifier, good quality parallel data were needed, as well as a dictionary that included translation probability. For this purpose, we used the TED talks [18] corpora. To obtain a dictionary, we trained a phrase table and extracted 1-grams from it [22].

## 5. Improvements to the mining process

Unfortunately, the native Yalign tool was not computationally feasible for large-scale parallel data mining. The standard implementation accepts plain text or web links, which need to be accepted, as input, and the classifier is loaded into memory for each pair alignment. In addition, the Yalign software is single-threaded. To make the process faster, a solution was developed that supplies the classifier with articles from the database within one session, with no need to reload the classifier each time. The developed solution also facilitated multi-threading and decreased the mining time by a factor of 6.1x (using a 4-core, 8-thread i7 CPU). The alignment algorithm was also reimplemented for better accuracy and to leverage the power of GPUs for additional computing requirements. The tuning algorithm was implemented as well.

### 5.1. Needleman-Wunsch algorithm (NW)

The objective of this algorithm is to align two sequences of elements (letters, words, phrases, etc.). The first step consists of defining the similarity between two elements. This is defined by the similarity matrix S, an NxM matrix, where N is the number of elements in the first sequence and M is the number of elements in the second sequence. The algorithm originated in the field of bioinformatics for RNA and DNA comparison. However, it can be adapted for text comparison. In simple terms, the algorithm associates a real number with each pair of elements in the matrix. The higher the number, the more similar the two elements are. For example, imagine that we have the similarity matrix S (phrase-polish, phrase-english) = number between 0 and 1. A 0 for two phrases means they have nothing in common; 1 means that those two phrases are the exact translation of each other. The similarity matrix definition is fundamental to the results of the algorithm [7].

The second step is the definition of the gap penalty. It is necessary in the case when one element of a sequence must be associated with a gap in the other sequence; however, such a step will incur a penalty (p).

The calculation of the S matrix is performed starting from the S(0,0) element that is, by definition, equal to 0. After the first row and columns are initialized, the algorithm iterates through the other elements of the S matrix, starting from the upper-left side to the bottom-right side. Each step of this calculation is shown in Figure 1.

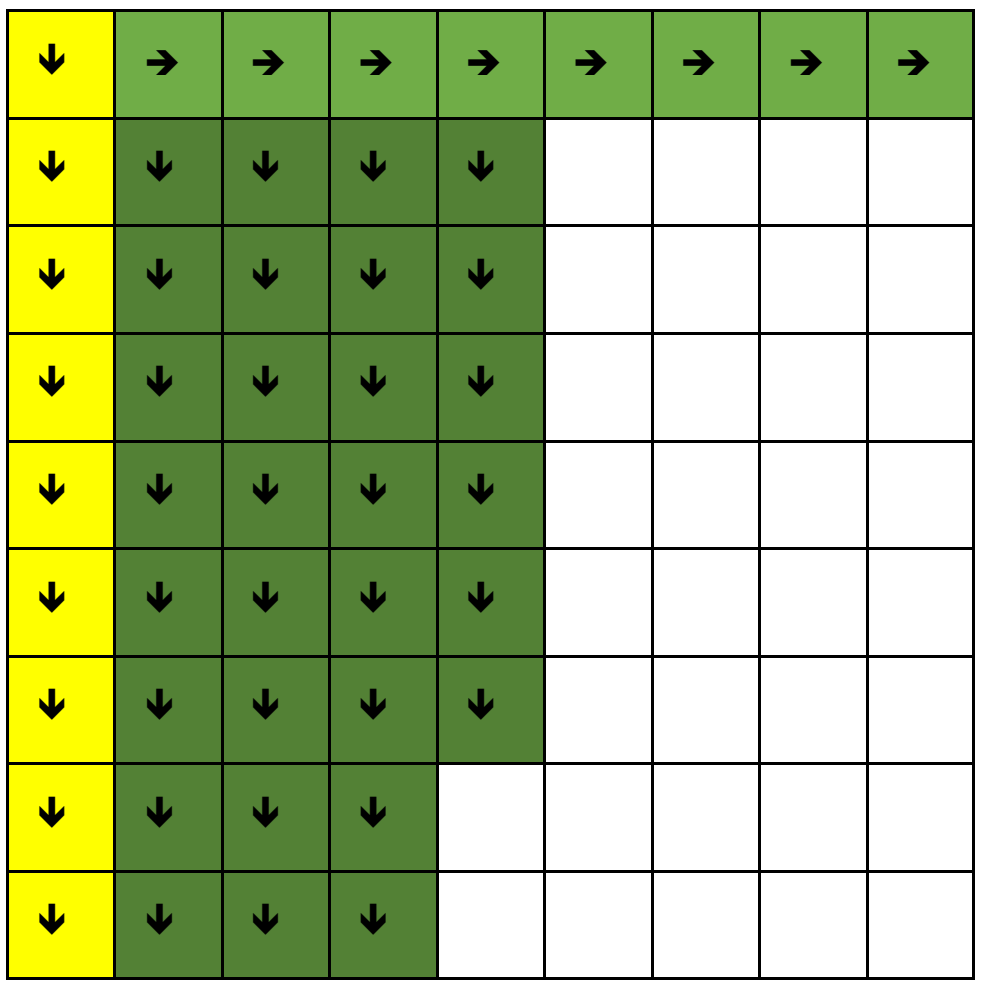

| | |
|---|---|
| ↓ (yellow) | First column initialization |
| → (light green) | First row initialization |
| ↓ (dark green) | Inside M matrix calculation |

*Figure 1:* Needleman-Wunsch S-matrix calculation

The two NW algorithms, with and without GPU optimization, are conceptually identical, but the first has an advantage in efficiency, depending on the hardware, of up to max(n, m) times.

It differs in the calculation of the S matrix elements. This calculation is the step to which multi-threading optimization is applied. Those operations are small enough to be processed by an enormous number of Graphics Processing Units (ex. CUDA cores). The idea is to compute all elements in a diagonal in parallel, always starting from the lower-left and proceeding to the bottom-right. An example is presented in Figure 2 [24].

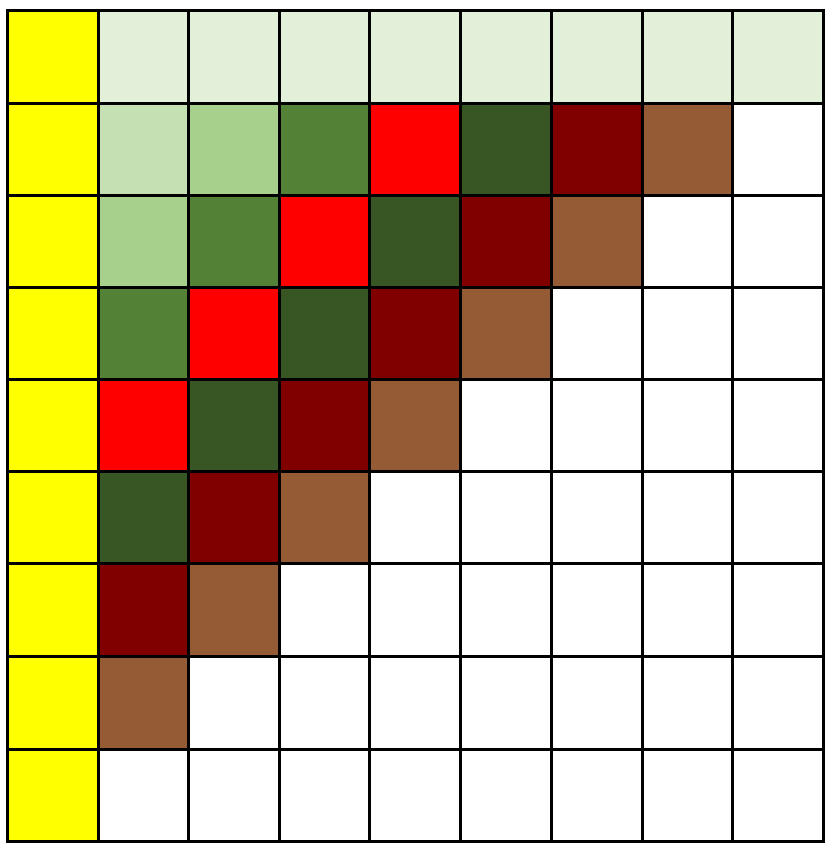

| | |
|---|---|
| (yellow) | First column initialization in parallel threads |
| (pale green) | First raw initialization in parallel |
| (light green) | 1st diagonal calculation in parallel threads |
| (green) | 2nd diagonal calculation in parallel threads |
| (dark green) | 3rd diagonal calculation in parallel threads |
| (red) | … |

*Figure 2:* Needleman-Wunsch S-matrix calculation with parallel threads

The S matrix calculation starts from the top left column. In order to find out the value of a cell of S(m,n), for all pairs of m and n, the values to its top S(m-1,n) , left S(m,n-1) and top left S(m-1,n-1) must be known in advance. Where, S(m,n) can be calculated with the help of following equation, [26]:

$$S(m,n) = \max\{S(m-1) \pm 1,\ S(m-1,n) - 2,\ S(m,n-1) - 2\} \quad (1)$$

Nonetheless, the results of the A* algorithm, if the similarity calculation and the gap penalty are defined as in the NW algorithm, will be the same only if there is an additional constraint on paths: paths cannot go upward or leftward in the S matrix. Yalign does not impose these additional conditions, so in some scenarios, repetitions of the same phrase may appear. In fact, every time the algorithm decides to move up or left, it is coming back into the second and first sequence respectively.

An example of an S matrix without constraints is presented in Figure 3:

| | a | d | e | g | f |
|---|---|---|---|---|---|
| a | X | | | | |
| d | | X | | | |
| c | X | | | | |
| d | | X | | | |
| e | | | X | X | X |

*Figure 3:* S matrix pass trough without constraints

The alignment result in this scenario is:
a, d, a, d, e, g, f
a, d, c, d, e, –, –

In the same problem, the NW would react as presented in Figure 4:

| | a | d | e | g | f |
|---|---|---|---|---|---|
| a | X | | | | |
| d | | X | | | |
| c | | X | | | |
| d | | X | | | |
| e | | | X | X | X |

*Figure 4:* S matrix pass trough with NW

The alignment result using NW would be:
a, d, –, –, e, g, f
a, d, c, d, e, –, –

In order to visualize the problem, let us assume that first sequence is “tablets make children very addicted” and second one is “tablets make people spoil children”. The solution to this sequence using A* algorithm without constrains is presented in Figure 5 and using NW in Figure 6.

| | tablets | make | children | very | addicted |
|---|---|---|---|---|---|
| tablets | X | - | - | - | - |
| make | - | X | - | - | - |
| people | X | - | - | - | - |
| spoil | - | - | - | - | - |
| children | - | - | X | X | X |

*Figure 5:A** alignment without constraints

| | tablets | make | children | very | addicted |
|---|---|---|---|---|---|
| tablets | X | - | - | - | - |
| make | - | X | - | - | - |
| people | - | - | - | - | - |
| spoil | - | - | - | - | - |
| children | - | - | X | - | - |

*Figure 6:* NW alignment with constraints

Because of the lack of constraints, repetitions were created that visualized the imperfection of the A* algorithm implemented in the Yalign program. Using A* many sentences may be misaligned or missed during the alignment, especially when analyzed texts are of different lengths and have vocabularies rich in synonyms. Some sentences can simply be skipped while checking for alignment. That is why NW with GPU optimization is more suitable algorithm. In this research, a comparison was made using all three approaches described above.

### 5.2. Tuning algorithm for classifier

The quality of alignments is defined by a tradeoff between precision and recall. The classifier has two configurable variables [25]:

- threshold: the confidence threshold to accept an alignment as "good." A lower value means more precision and less recall. The "confidence" is a probability estimated from a support vector machine classifying "is a translation" or "is not a translation." [27]
- penalty: controls the amount of "skipping ahead" allowed in the alignment [5]. Say you are aligning subtitles, where there are few or no extra paragraphs and the alignment should be more or less one-to-one; then the penalty should be high. If you are aligning things that are moderately good translations of each other, where there are some extra paragraphs for each language, then the penalty should be lower.

Both of these parameters are selected automatically during training but they can be manually adjusted if necessary. The solution implemented in this research also introduces a tuning algorithm for those parameters, which allows for better adjustment of them.

To perform tuning, it is necessary to extract random article samples from the corpus. Such articles must be manually aligned by humans. Based on such information, the tuning script tries, naively by random parameter selection, to find values for which classifier output is as similar to that of a human as possible. Similarity is a percentage value of how the automatically-aligned file resembles the human-aligned one. A Needleman-Wunsch algorithm is used for this comparison. Analysis was performed for each of four languages of interest to check how the tuning algorithms cope with proper adjustment of the parameters. Table 1 shows the results of this experiment. For testing purposes, 100 random article pairs were taken from the Wikipedia comparable corpus and aligned by a human translator. Second, a tuning script was run using classifiers trained on the previously described text domain. A percentage change in quantity of recognized parallel sentences was calculated for each classifier.

*Table 1*: Improvements in mining using tuning script for Wikipedia data

| Domain | Improvement in % |
|---|---|
| DE | 11.2 |
| FR | 13.5 |
| CS | 12.1 |
| VI | 15.2 |

### 5.3. Minor improvements for better Wikipedia exploration

All the improvements discussed in the previous sections deal mostly with heuristics used in the mining tool and can be applied to any bilingual textual data. Such an attitude would also improve the mining within Wikipedia. However, Wikipedia has many additional sources of cross-lingual dependencies that can be used. First of all, the topic domain of Wikipedia cannot be closed into a specific domain; this page covers almost any topic. This is the reason why its articles often contain complicated or rare vocabulary, and why statistical mining methods may skip some parallel sentences. The solution to this problem might be extraction of the dictionary using the article titles from Wikipedia (Figure 7) and additionally implemented web crawler tool.

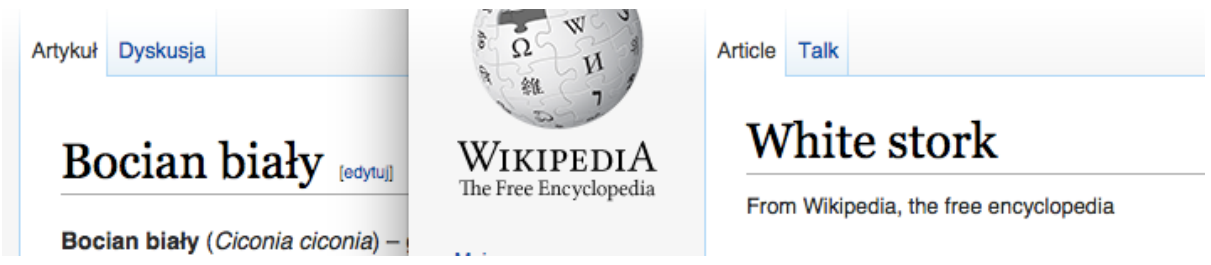

*Figure 7:* Example of bi-lingual Wikipedia page title

According to [11] precision of such a dictionary can be as high as 92%. Such a dictionary can be used not only for the extension of the parallel corpora but also in the classifier training phase.

Secondly, figures, or to be more precise, their descriptions, contain good quality parallel phrases. It is possible to get them by picture analysis and hyperlinks to the pictures. The same goes for any figures, tables, maps, audio, video or any other multimedia contents on Wikipedia. Unfortunately, not all information can be extracted from Wikipedia dumps and it is required to use a web crawler suited for this task (Figure 8). This also means that only cross-lingual information that are annotated with common links can be extracted.

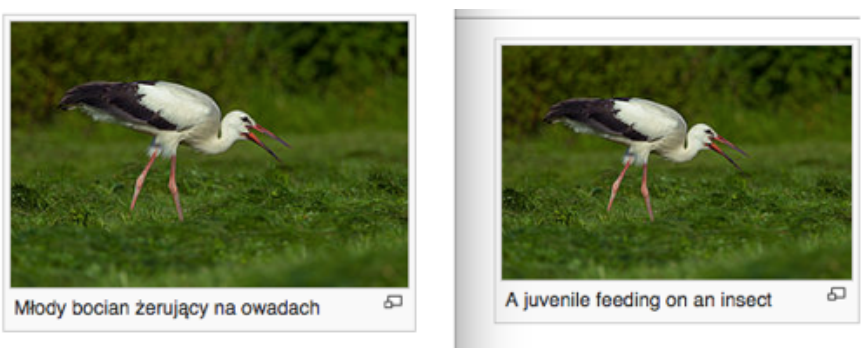

*Figure 8:* Example of bi-lingual figure caption

The good quality Wikipedia articles are well referenced. It is most likely for sentences to be cross-lingual equivalents if they are referenced with the same publication. Such analysis, joined with other comparison techniques, can lead to better accuracy in parallel text recognition (Figure 9).

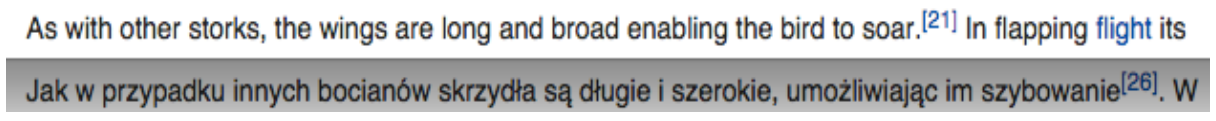

*Figure 9:* Example of bilingually referenced sentence

Unfortunately, the Wikipedia articles are developed separately for each language by many authors. In the following example, the parallel sentence in one language is notated with reference number 21 and in other with reference 26. It is why it is required not only to compare the number but to analyze the references themselves (Figure 10).

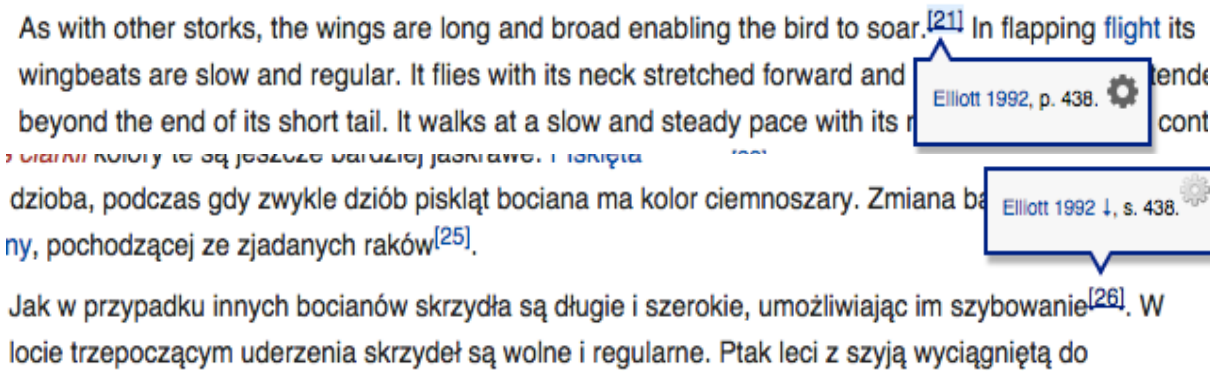

*Figure 10.* Example of cross-lingual reference

In addition to references, it is important to analyze names, dates, numbers etc. as well, because it they indicate parallel data presence.

Because of the need to use a web crawler, this tool version was evaluated only using 1,000 randomly selected articles. It would take too long to build using an entire corpus without access to many proxy servers and Internet connections. It is not only required to crawl about 5,000,000 articles for EN wiki but also many links present in each language Wiki.

*Table 2:* Number of parallel segments found

| | |
|---|---|
| Y | 4,192 |
| YMOD | 5,289 |
| DICT | 868 |
| DICTC | 685 |

Firstly, the data was crawled and secondly aligned using the standard version of the classifier (Y in Table 2) and then aligned using the modified version - dictionary, captions and references extraction as described above (YMOD in Table 2). Lastly, the single words were extracted and counted (DICT in Table 2) and also manually analyzed in order to verify how many of them could be considered as correct translations (DICTC in Table 2).

The results mean that the improved method using additional information sources mined an additional 1,097 parallel segments. Out of them, it possible to identify 868 single words, which means that in fact 229 new sentences were obtained. Potential growth in obtained data was equal to 5.5%. After manual analysis of the dictionary, 685 words were identified as proper translations. This means that the accuracy of the dictionary was about 79%.

### 5.4. Evaluation of improvements

As mentioned, some methods for improving the performance of the native classifier were developed. First, speed improvements were made by introducing multi-threading to the algorithm, using a database instead of plain text files or Internet links, and using GPU acceleration in sequence comparison. More importantly, two improvements were obtained to the quality and quantity of the mined data. The A* search algorithm was modified to use Needleman-Wunch, and a tuning script of mining parameters was developed. In this section, the CS-EN TED corpus will be used to demonstrate the impact of the improvements (it was the only classifier used in the mining phase). The data mining approaches used were: directional (CS->EN classifier) mining (MONO), bi-directional (additional EN->CS classifier) mining (BI), bi-directional mining using a GPU-accelerated version of the Needleman-Wunch algorithm (NW), and mining using the NW version of the classifier that was tuned (NWT). Such mining methodologies were already successfully evaluated against MT quality in [25]. The results of such mining are shown in Table 3.

*Table 3*: Number of obtained Bi-Sentences

| Mining Method | Number of Bi-Sentences |
|---|---|
| MONO | 21,132 |
| BI | 23,480 |
| NW | 24,731 |
| NWT | 27,723 |

As presented in Table 3, each of the improvements increased the number of parallel sentences discovered. In addition, in Table 4 a speed comparison is made using different versions of the tool.

*Table 4*: Computation Time of Different Yalign Version

| Mining Method | Computation Time [s] |
|---|---|
| Y | 92.37 |
| MY | 15.1 |
| NWMY | 18.2 |
| GNWMY | 16.4 |

A total of 1,000 comparable articles were randomly selected from Wikipedia and aligned using the native implementation (Y), multi-threaded implementation (MY), classifier with the Needleman-Wunch algorithm (NWMY), and with a GPU-accelerated Needleman-Wunch algorithm (GNWMY)

The results indicate that multi-threading significantly improved speed, which is very important for large-scale mining. As anticipated, the Needleman-Wunch algorithm decreases speed. However, GPU acceleration makes it possible to obtain performance almost as fast as that of the multi-threaded A* version. It must be noted that the mining time may significantly differ when the alignment matrix is big (text is long). The experiments were conducted on a hyper-threaded Intel Core i7 CPU and a GeForce GTX 660 GPU.

## 6. Evaluation of obtained comparable corpora

Using techniques described above, we were able to build comparable corpora and mine them for parallel sentences for the four languages being part of IWSLT 2015 evaluation campaign. We used GPU accelerated Needleman-Wunch algorithm, the classifier was tuned and Wikipedia page titles were downloaded separately. We focused on DE, FR, CS and VI. The corpora statistics are presented in Table 5.

*Table 5:* Results of mining after improvements

| Language Pair | Number of bi-sentences | Number of unique EN tokens | Number of unique foreign tokens |
|---|---|---|---|
| DE-EN | 2,459,662 | 2,576,938 | 2,864,554 |
| FR-EN | 818,300 | 1,290,000 | 1,120,166 |
| CS-EN | 27,723 | 98,786 | 104,596 |
| VI-EN | 58,166 | 92,434 | 93,187 |

To evaluate the corpora, we trained baseline systems using IWSLT 2015 official data sets and enriched them with obtained comparable corpora, both as parallel data and as language models. The enriched systems were trained with the baseline settings but additional data was adapted using linear interpolation and Modified Moore-Lewis [23]. Because of the well know MERT instability, tuning was not performed in the experiments [20]. Using MERT would most likely improve overall MT systems quality but it some cases it could produce false positive results, what needed to be avoided in order to properly evaluate only the impact of augmented corpora [20].

*Table 6:* Results of MT Experiments

| LANGUAGE | SYSTEM | DIRECTION | BLEU |
|---|---|---|---|
| DE-EN | BASE | →EN | 30.21 |
| | EXT | →EN | 31.37 |
| | BASE | EN← | 21.07 |
| | EXT | EN← | 22.47 |
| FR-EN | BASE | →EN | 35.95 |
| | EXT | →EN | 37.01 |
| | BASE | EN← | 35.73 |
| | EXT | EN← | 37.79 |
| CS-EN | BASE | →EN | 23.55 |
| | EXT | →EN | 24.09 |
| | BASE | EN← | 14.05 |
| | EXT | EN← | 14.93 |
| VI-EN | BASE | →EN | 22.84 |
| | EXT | →EN | 23.38 |
| | BASE | EN← | 26.23 |
| | EXT | EN← | 26.76 |

The evaluation was conducted using official test sets from IWSLT 2010-2013 campaigns and averaged. For scoring purposes, Bilingual Evaluation Understudy (BLEU) metric was used. The results of the experiments are shown in Table 6. BASE in the Table 6 stands for baseline system and EXT for enriched systems.

As anticipated, additional data sets improved overall translation quality for each language and in both translation directions. The gain in quality was observed mostly in the English to foreign language direction.

## 7. Conclusions

Bi-sentence extraction has become more and more popular in unsupervised learning for numerous specific tasks. This method overcomes disparities between English and other languages. It is a language-independent method that can easily be adjusted to a new environment, and it only requires parallel corpora for initial training. Our experiments show that the method performs well. The resulting corpora increased MT quality in a wide text domain. In some cases, only very small BLEU score differences were reported. Nonetheless, it can be assumed that even small differences can make a positive influence on real-life, rare translation scenarios. In addition, it was proven that mining data using two classifiers trained from a foreign to a native language and vice versa, can significantly improve data quantity, even though some repetitions are possible. From a practical point of view, the method requires neither expensive training nor language-specific grammatical resources, but it produces satisfying results. It is possible to replicate such mining for any language pair or text domain, or for any reasonable comparable input data.

## 8. Acknowledgements

This work is supported by the European Community from the European Social Fund within the Interkadra project UDA-POKL-04.01.01-00-014/10-00 and PJATK statutory resources ST/MUL/02/2015.